\documentclass[10pt, a4paper]{article}

\usepackage[final]{lrec2026} 

\newcommand{\dataset}[0]{\textsc{RedHOTExpect}} 
\newcommand{\redhot}[0]{\textsc{RedHOT}} 

\usepackage{listings}
\usepackage{amsmath}
\usepackage{cleveref} 
\crefname{section}{\S}{\S\S}
\Crefname{section}{\S}{\S\S}
\crefname{table}{Table}{Tables}
\crefname{figure}{Figure}{Figures}
\crefname{algorithm}{Algorithm}{}
\crefname{equation}{eq.}{}
\crefname{appendix}{App.}{}
\crefformat{section}{\S#2#1#3} 
\crefformat{subsection}{\S#2#1#3}
\crefformat{subsubsection}{\S#2#1#3}
\usepackage{annotates}
\usepackage[dvipsnames,svgnames,x11names]{xcolor}
\definecolor{myPurple}{HTML}{B266FF}
\definecolor{myMaroon}{HTML}{9B2C2C}
\usepackage{todonotes}
\usepackage{booktabs}
\makeatletter
\newcommand*\iftodonotes{\if@todonotes@disabled\expandafter\@secondoftwo\else\expandafter\@firstoftwo\fi} 
\makeatother

\usepackage{tabularx}

\renewcommand{\paragraph}[1]{\par\noindent\textbf{#1}}

\providecommentcommand{amw}{myPurple}{Amelie}
\providecommentcommand{asw}{myMaroon}{Aswathy}
\usepackage{paralist}

\usepackage{fvextra} 
\DefineVerbatimEnvironment{promptblock}{Verbatim}
{fontsize=\small,breaklines=true,breakanywhere=true}

\title{Towards Expectation Detection in Language: A Case Study on Treatment Expectations in Reddit}

\name{Aswathy Velutharambath, Amelie W\"uhrl} 

\address{University of Stuttgart, Germany, IT University of Copenhagen, Denmark \\
         aswathy.velutharambath@ims.uni-stuttgart.de, amwy@itu.dk\\}

\abstract{
Patients' expectations towards their treatment have a substantial effect on the treatments' success. While primarily studied in clinical settings, online patient platforms like medical subreddits may hold complementary insights: treatment expectations that patients feel unnecessary or uncomfortable to share elsewhere. 
Despite this, no studies examine what type of expectations users discuss online and how they express them. Presumably this is because expectations have not been studied in natural language processing (NLP) before. Therefore, we introduce the task of Expectation Detection, arguing that expectations are relevant for many applications, including opinion mining and product design. 
Subsequently, we present a case study for the medical domain, where expectations are particularly crucial to extract. 
We contribute RedHOTExpect, a corpus of Reddit posts (4.5K posts) to study expectations in this context. We use a large language model (LLM) to silver-label the data and validate its quality manually (label accuracy $\approx$78\%). Based on this, we analyze which linguistic patterns characterize expectations and explore what patients expect and why.
We find that optimism and proactive framing are more pronounced in posts about physical or treatment-related illnesses compared to mental-health contexts, and that in our dataset, patients mostly discuss benefits rather than negative outcomes. The RedHOTExpect corpus can be obtained from  https://www.ims.uni-stuttgart.de/data/RedHOTExpect
\\
 }

\begin{document}

\maketitleabstract

\section{Introduction}
Expressions of expectations are ubiquitous in natural language. They hold valuable information to monitor public opinion toward policy changes (see Table~\ref{tab:expectation-examples}, Ex.~1), detect student needs (Ex.~2), or inform product design (Ex.~3).  
In the medical domain they are particularly relevant:
Expectations toward a treatment are powerful psychosomatic mechanisms known to affect treatment success~\cite{bingel2011effect, rief2016rethinking, aulenkamp2023nocebo}. Pain medication, for example, can be twice as effective if the patient expects the treatment to work, while expecting it not to work can cancel its effect~\cite{bingel2011effect}.
So far, studying these effects relies on small-scale patient surveys. While indispensable, their scale is limited, which leads to the following constraints: they might fail to detect expectations that patients do not consider `report-worthy', an issue that~\citet{duh-et-al_2016} show for reports of adverse drug effects. Additionally, surveys are known to be biased, leaving knowledge gaps for underrepresented groups~\cite{weber-et-la_2021}. 
Both hinder doctors from addressing concerns caused by negative expectations and to achieve effective treatment.

\begin{table}[]
    \centering
    \begin{tabularx}{\columnwidth}{cX}
    \toprule
        id& Texts expressing expectation \\
        \cmidrule(r){1-1} \cmidrule(l){2-2}
        1 & All this talk about the new bus lanes. At least I'll have a new excuse for being late to work from now on.\\
        2 & I'm pretty optimistic that the new exam format might finally test how well we think, not how long we can sit still\\
        3 & This phone should last at least three years\\
        4 & How is this supposed to have an effect with all the weed I've been smoking?\\
        5 & It's a stupid reason, but everyone says how the meds can make you feel bloated.\\
        6 & I bet no one tested this for darker skin tones. Makes me scared to even fill the prescription\\
        \bottomrule
    \end{tabularx}
    \caption{Example texts expressing expectation across general and medical domain.}
    \label{tab:expectation-examples}
\end{table}

Social media, where patients frequently discuss medical experiences, enables us to study expectations and their effects in `real-world' environments. 
Here, patients may share skepticism towards a treatment, even if they are uncomfortable or find it unnecessary to disclose their concerns with medical authorities~(see Ex.~4 and 5).
Similarly, marginalized groups who have experienced medical discrimination may distrust doctors' recommendations~(see Ex.~6) and therefore discuss treatment expectations and concerns mostly online. Identifying such online discussions can help practitioners anticipate a wider range of concerns that may foster negative expectations.

However, we are lacking a crucial prerequisite to extract this knowledge: research on detecting expectations from text. 
While tasks like hope and regret detection~\cite{chakravarthi-2020-hopeedi, BALOUCHZAHI2023120099, balouchzahi2023polyhope, balouchzahi-et-al-2024_hope-regret-X},  or complaint detection~\cite{jin-aletras-2020-complaint} are related to expectations, the linguistic characteristics and domain-specific properties of expectations are not studied. 
We assume this is because identifying descriptions of expectations is difficult. As illustrated by the examples in Table~\ref{tab:expectation-examples}, expectations are often implicit, nuanced, or require understanding contextual cues. Current LLMs may be capable of extracting them. To the best of our knowledge, however, they have not been evaluated for this task. This is likely because we still lack a comprehensive definition of relevant expectation types and resources for evaluation.

Therefore, we make the following contributions:

(1) We introduce \textit{Expectation Detection} as a novel NLP task. The goal is to identify linguistic and stylistic patterns associated with expression of \emph{expectation} in which a writer conveys a belief, anticipation, or prediction about a possible future outcome. Unlike sentiment or stance analysis, which evaluate attitudes about the present, past or future, expectation detection specifically targets \emph{future-oriented mental states}: utterances in which a person articulates what they think, hope, or fear may happen. 

(2) Considering the crucial impact expectations have on treatment success, we start exploring this task with a case study on patient-authored health texts. 
To this end, we contribute the first resource to study treatment expectations in social media text. 
The corpus, \dataset, consists of $\approx$4.5K medical Reddit posts expressing expectation, of which around 2.5K include token-level annotations for treatment, expectations, and outcome descriptions (TEO triplets). Hypothesizing that LLMs have a rudimentary notion of `expectation', we use an LLM to filter and silver-label expectations and outcomes. We manually validate the LLM annotations in a representative subset of the data (250 instances), and find them to be accurate in 77.5\% of cases.
For the same subset, we additionally characterize expectation and outcome mentions with respect to their type and the basis of expectation. 

(3) Using this data, we study \emph{how patients discuss treatment expectations on social media} by analyzing which linguistic and stylistic patterns characterize these utterances~(\Cref{sec:rq1}). In a targeted manual analysis~(\Cref{sec:rq2}, we explore \emph{what patients expect and why}.
We find that expectation-related discourse is characterized by a more optimistic and action-oriented tone, whereas non-expectation posts more often convey caution, frustration, or negative affect.
Overall, we observe that optimism and proactive framing are most pronounced in posts about physical or treatment-related illnesses, whereas mental-health contexts convey expectations in a more introspective and socially detached tone.
Notably, expectations in our dataset mostly discuss benefits rather than negative outcomes.

\section{Related Work}
\label{sec:related-work}

Patients frequently use online platforms to discuss medical topics~\cite{chen2021social-media-health-review}. Therefore, a subfield of medical NLP specializes on biomedical content from social media and other online sources. Central tasks are pharmacovigilance, i.e., identifying descriptions of adverse drug reactions~\citep{nikfarjam-et-al_2015,Cocos2017,magge-et-al_2021,karimiet-al_2015}, monitoring public health~\citep{Paul2012,Choudhury2013,Sarker2016,stefanidis-et-al_2017}, extracting personal health experiences~\citep{Yin2015,klein-et-al_2017,Karisani2018,wadhwa-etal-2023-redhot, falk-lapesa-2024-stories}, analyzing mental health discourse~\citep{garg_mental_2023}, and fact-checking~\citep[i.a.]{hossain-et-al_2020,mattern-et-al_2021,saakyan-et-al_2021,kim-etal-2023-covid}.
%
Health mining datasets are commonly collected from patient fora~\cite{karimiet-al_2015, vladika-etal-2024-healthfc} and various social media platforms. Until the company limited its API access, Twitter (now X) served as a popular data source~\cite[i.a.]{Sarker2016, wuhrl-klinger-2022-bear, sundriyal-etal-2022-empowering}. Reddit is another frequently used source, particularly because posts are not length-restricted ~\cite[i.a.]{scepanovic-et-at_2020,basaldella-et-al_2020,wadhwa-etal-2023-redhot}. Recently, synthetic data also has become increasingly popular~\cite[i.a.]{wuehrl-etal-2024-ims-medically, yamagishi-nakamura-2024-utrad, Ghanadian2024SociallyAS}.

While connected to concepts such as hope~\citep{balouchzahi2023polyhope}, stance~\cite{hardalov-etal-2022-survey}, or emotion~\cite{plaza-del-arco-etal-2024-emotion}, which have been studied in NLP, expectation detection is an unexplored task. 
Some work addresses adjacent tasks: \citet{jin-aletras-2020-complaint} study complaint detection, which involves identifying when someone voices a mismatch between their expectation and reality. Similarly, work on conversational intent~\cite{sakurai-miyao-2024-intention-detection-llms} may capture the speakers' expectation of how the conversation could evolve. 
The most closely related is work on hope and regret detection~\cite{chakravarthi-2020-hopeedi, BALOUCHZAHI2023120099, balouchzahi2023polyhope, balouchzahi-et-al-2024_hope-regret-X}. While an expression of hope conveys an expectation, it is limited to a positive stance towards the event. Regret always expresses a negative attitude towards a past event.
\citet{bringula2022students-expectations} extract expectations in student essays using word clustering and sentiment counts, but do not model expectation expressions or their linguistic grounding.

Some work brings together concepts relevant to expectations with health tasks. With respect to emotion, \cite{khanpour-caragea-2018-Emotion-Detection-Health-Related-Online-Posts} detect patients' emotional states from online posts.
\cite{turcan-etal-2021-emotion-psychological-stress} detect psychological stress using emotion-informed models, while others explore understanding emotional states towards medical events like pandemics~\citep[i.a.]{ng-etal-2020--emotion-covid, sosea-etal-2022-emotion-covid}.
The work most related to our case study on extracting treatment expectations is~\citet{balouchzahi-et-al-2024_hope-regret-X} who model expressions of hope and regret with respect to drug use, and 
\citet{ettlin2025patients} who explore patient expectations in clinical self-reports using topic modeling and linguistic style analysis 

To the best of our knowledge, we are the first to formalize the task of expectation detection in NLP. So far, no work on extracting treatment expectations from social media exist. This motivates creating a resource that allows us to study how patients discuss their expectations towards a medical treatment.

\section{Expectation Detection}
\label{sec:expectation-detection}
We formalize the task of \textit{Expectation Detection} by defining what constitutes an expectation and what linguistic properties can be attributed to it.

\paragraph{Defining Expectation Event.}
We define an \textit{Expectation Event} (EE) as any statement or implication about a possible or likely future state, whether beneficial, harmful, or neutral, that is explicitly linked to an action, intervention, or situation. In linguistic terms, EEs capture how people verbalize their belief, hope, or fear about a \textit{future} outcome (See \Cref{tab:expectation-examples}).  
To establish what constitutes an expectation and how it differs from related constructs such as sentiment or stance, we attribute the following properties to each Expectation Event:

\begin{compactitem}
	\item \textbf{Expectation Type:} the valence or direction of the anticipated outcome, such as \emph{benefit}, \emph{harm}, \emph{no effect}, \emph{worsening} or \emph{mixed}, or \emph{other}.
	
	\item \textbf{Expectation Basis:} the source or justification underlying the belief, e.g., \emph{Personal} (based on one's own past experience),  
	\emph{Social} (based on peers or community experiences),  
	\emph{Authority} (based on statements from doctors or experts),  
	\emph{Information/Media} (based on studies, advertisements, or online information),  
	\emph{Cultural} (stemming from social or cultural beliefs),  
	\emph{Self-efficacy} (based on belief in one's ability to follow the treatment).
	
	\item \textbf{Certainty:} the degree of confidence or tentativeness expressed by the author (e.g., ``\textit{I hope it helps}'' vs. ``\textit{It will definitely help}'').
	\item \textbf{Temporal Orientation:} whether the expectation is \emph{prospective} (before the outcome is known) or \emph{retrospective} (recalled after an outcome has occurred).
\end{compactitem}

\Cref{fig:ee-example} shows examples of expectations annotated with these dimensions. 

\paragraph{Task Definition.}
This study examines expectations expressed in medical discourse.  
We specifically target \textbf{outcome-related expectations}, statements that express anticipated effects of a treatment or intervention that could plausibly lead to a health-related outcome.  

For example, ``\textit{I think this new medication will finally help me sleep}'' and ``\textit{I'm worried the drug might make me nauseous}'' both describe future outcomes linked to a treatment and thus constitute Expectation Events.  
By contrast, ``\textit{I expect to see my doctor next week}'' expresses an expectation but not one about a treatment outcome and therefore falls outside our scope.  

Formally, given a text span \(T\), the Expectation Detection task in this study involves two subtasks:  
\begin{compactenum}
\item \textbf{Expectation Identification}: determine whether \(T\) contains one or more Expectation Events; and  
\item \textbf{Treatment-Expectation-Outcome annotation}: if so, identify their key components: (a) the \emph{treatment or intervention} about which the expectation is expressed), (b) the \emph{expectation} (what is expected to occur) and (c) the \emph{actual outcome} (what actually occurred).
\end{compactenum}

\begin{figure}[t]
	\centering
	\fbox{
		\begin{minipage}{0.95\linewidth}
			\small
			\textbf{Example 1:}
			\textit{``My doctor prescribed me X. I really hope it will stop my headaches this time, but I'm a bit afraid it could make me dizzy.''} \\[4pt]
			\textbf{Expectation Events:}
			\begin{itemize}
				\item[\textbf{EE1}] \textit{``I really hope it will stop my headaches''} \\
				\quad \textbf{Expectation Type:} Benefit \\
				\quad \textbf{Expectation Basis:} Authority (doctor's prescription) \\
				\quad \textbf{Certainty:} Moderate (``hope'') \\
				\quad \textbf{Temporal Orientation:} Prospective
				\item[\textbf{EE2}] \textit{``I'm a bit afraid it could make me dizzy.''} \\
				\quad \textbf{Expectation Type:} Harm \\
				\quad \textbf{Expectation Basis:} None \\
				\quad \textbf{Certainty:} Low (``could'') \\
				\quad \textbf{Temporal Orientation:} Prospective
			\end{itemize}
            \textbf{Example 2:}
            \textit{``All this talk about the new bus lanes. At least I'll have a new excuse for being late to work from now on''} \\[4pt]
			\textbf{Expectation Events:}
			\begin{itemize}
				\item[\textbf{EE1}] \textit{bus lanes will lead to delays in traffic} \\
				\quad \textbf{Expectation Type:} Worsening \\
				\quad \textbf{Expectation Basis:} Information/Media \\
				\quad \textbf{Certainty:} High \\
				\quad \textbf{Temporal Orientation:} Prospective

			\end{itemize}
		\end{minipage}
	}
	\caption{Annotated examples illustrating Expectation Events (EEs) in posts.}
	\label{fig:ee-example}
\end{figure}

\section{Expectation Corpus}

\begin{figure*}
	\centering
	\includegraphics[width=0.9\linewidth]{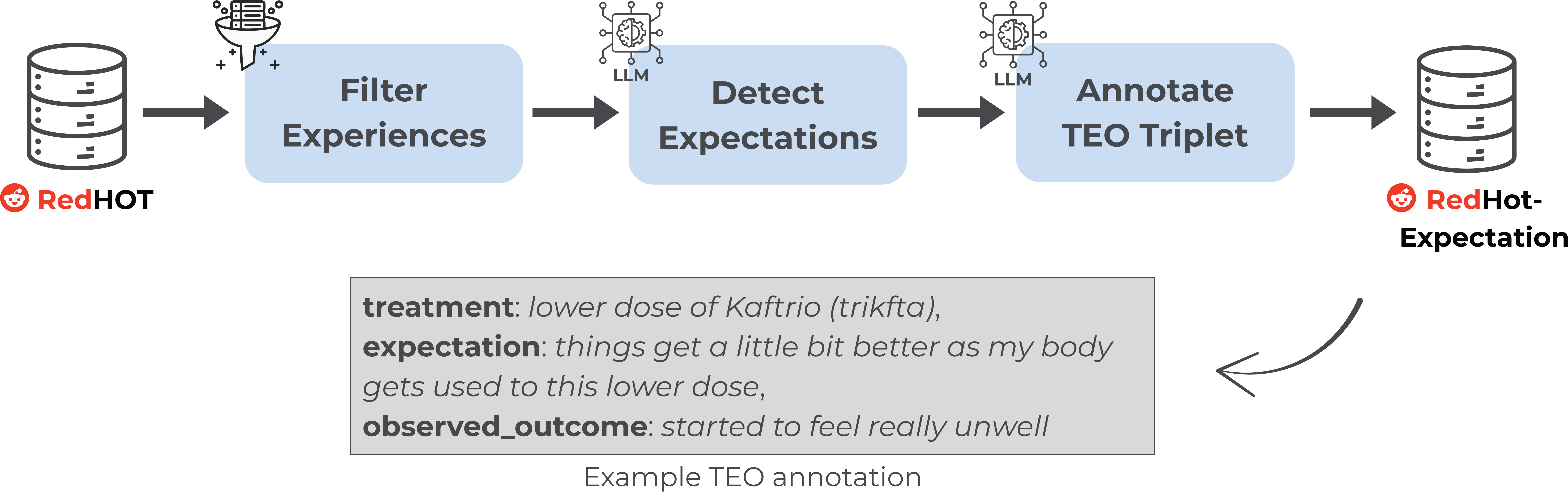}
	\caption{Automated pipeline for creating annotated RedHOT-Expectation data.}
	\label{fig:pipeline}
\end{figure*}

\subsection{Corpus Creation}
\label{sec:corpus-creation}

To systematically study how patients express expectations about medical treatments on social media, we  require data that captures this concept. To facilitate this, we construct \dataset, a corpus of Reddit posts containing expressions of expectation, further annotated with Treatment–Expectation–Outcome (TEO) triplets. The following section outlines the steps involved in creating the corpus. The overall annotation process is illustrated in \Cref{fig:pipeline}.

\paragraph{Experience filtering.}
As a starting point, we build on the existing \textsc{RedHOT} corpus~\cite{wadhwa-etal-2023-redhot}, which contains $\approx$22k Reddit posts related to health conditions. We selected this dataset because it already focuses on patient-authored health discussions and provides high-level discourse annotations such as \emph{questions}, \emph{experiences}, and \emph{claims}. This structure allows us to efficiently identify posts in which users describe their own medical experiences, which are the contexts where treatment expectations are most likely to appear. We therefore subset all posts labeled as \emph{experience}, resulting in a collection of $\approx$12k posts. This filtered subset forms the foundation for identifying explicit mentions of expectations in the following step.

\paragraph{Expectation labeling.}
Identifying expressions of expectation in Reddit posts is challenging because the texts are often long, narrative, and loosely structured. Expectations are frequently implicit, context-dependent, and not always expressed within a continuous span of text. For instance, a post may introduce a medication or treatment at the beginning and describe the anticipated outcome only at a later stage. Given this complexity, we employ a large language model to leverage its broad semantic understanding and contextual reasoning capabilities for identifying posts that express expectations. Specifically, we use the open-weight model \texttt{gpt-oss-20b}\footnote{\url{https://huggingface.co/openai/gpt-oss-20b}} to classify each post as either \emph{contains expectation} or \emph{not}. The model is prompted with a detailed definition of outcome-related expectations and few-shot examples illustrating positive and negative cases. It is based on the expectation definition and examples discussed in \cref{sec:expectation-detection}\footnote{See Appendix~\ref{sec:appendix-prompt}for the full prompt.}. We conduct several iterations of prompt refinement and manual verification to optimize the precision expectation detection.

\paragraph{Treatment-Expectation-Outcome triplet.}
In addition to examining how expectations are expressed linguistically, a further motivation for building the \dataset\ corpus is to enable analyses of how treatment expectations relate to reported outcomes. To support this goal, we extend the annotation to a structured \emph{Treatment–Expectation–Outcome (TEO)} triplet, where the \emph{Treatment} refers to the intervention or medication mentioned, the \emph{Expectation} to the anticipated outcome, and the \emph{Outcome} to the described or implied result. We again use the open-weight model \texttt{gpt-oss-20b}, providing detailed prompts to automatically generate these triplets.

Given the narrative style of Reddit posts, we instruct the model to extract one or more TEO triplets per post, typically one per treatment mention. Because Reddit threads often include exchanges between community members, an author may report the treatment outcome only later in the discussion. To account for this, we retrieve all comments posted by the same author and append them to the prompt, enabling the model to detect outcomes described in follow-up messages. This procedure allows us to capture both \emph{prospective} expectations (before the outcome is known) and \emph{retrospective} expectations (beliefs recalled after an outcome has occurred).

\begin{table}
	\centering
	\resizebox{\columnwidth}{!}{%
		\begin{tabular}{lrrrrr}
			\cmidrule{1-6}
			\textbf{Class / Subset} & \textbf{Count} & \textbf{Mean} & \textbf{Std} & \textbf{Median} & \textbf{Max} \\
			\cmidrule(lr){1-6}
			No expectation \quad & 7,364 & 136.6 & 124.7 & 102 & 1,771 \\
			Expectation (all) \quad & 4,433 & 182.9 & 170.8 & 138 & 2,944 \\[2pt]
			\cmidrule{1-6}
			\quad has TEO \quad & 2,529 & 212.6 & 198.7 & 158 & 2,944 \\
			\quad no TEO \quad & 1,904 & 143.6 & 113.0 & 113 & 1,042 \\
			\hline
		\end{tabular}
	}
	\caption{Token length statistics by class and within the subset of expectation posts, distinguishing those with and without extracted treatment–expectation–outcome (TEO) triplets.}
	\label{tab:length-stats}
\end{table}

\subsection{Human Validation}\label{subsec:human}
To evaluate how reliably the LLM extracts treatment--expectation--outcome (TEO) triplets we annotate a subset
of instances manually. We select a stratified sample of 245 posts\footnote{Originally sampled 250, five exclude because expectations were labeled "None" or "null".} from the TEO annotated subset of the \dataset. To ensure diversity while maintaining relevance, we restrict the sample to condition--treatment pairs that occur at least three times in the corpus, 
guaranteeing sufficient contextual coverage across domains. The sampling 
procedure includes at least one instance of each eligible pair, with the 
remaining posts selected at random and balanced across health conditions.

Out of the 245 manually inspected posts, the automatic pipeline had extracted 502 Treatment--Expectation--Outcome (TEO) triplets. Manual validation confirms that 389 of these are correct, distributed across 200 posts, while 45 posts contained no valid triplet. This corresponds to a labeling accuracy of 77.5\%, indicating that the automatic extraction process achieves a reasonably high level of precision while allowing for multiple expectations to be captured within a single post.

\subsection{Corpus Details}
The initial subset of posts drawn from the \redhot dataset contained 11,797 Reddit posts covering 23 health conditions. Each post was automatically classified by a large language model as either containing or not containing an expectation-related expression, following the procedure described in \Cref{sec:corpus-creation}. Overall, 4,433 posts (37.6\%) were labeled as containing treatment expectations, suggesting that expectation-related discourse is a frequent phenomenon in patient-authored online discussions. The remaining 7,364 posts (62.4\%) describe treatment experiences, advice, or symptom trajectories without explicitly expressing anticipation or belief about a future treatment outcome. 
In the TEO annotation step, out of the 4,433 posts labeled as \textit{contains expectation}, 2,529 posts (57.0\%) contained at least one complete triplet linking a treatment, an expectation, and an outcome. 

From the manual validation step (\Cref{subsec:human}) we observe an identification accuracy of approximately 78\%, implying an expected error rate of around 22\%. 
For the subsequent analyses, we should keep this in mind and interpret findings in light of the estimated labeling noise.

\section{How do patients discuss treatment expectations on social media?}
\label{sec:rq1}
Our goal is to identify how language use differs between posts that express treatment expectations and other health-related narratives.
We focus on linguistic and stylistic markers that may capture how people formulate predictions, intentions, or outcome-oriented thinking in health discourse.

\subsection{Experimental setup}
To examine linguistic markers of expectation in health-related discourse, we perform a lexical feature analysis using the \textsc{LIWC-22} dictionary~\citep{boyd2022liwc22}. It provides frequency-based indicators of psychological and linguistic processes and is widely used in NLP and social media text analysis.
In addition, we include sentiment analysis using the VADER lexicon~\citep{Hutto_Gilbert_2014}, the Gunning--Fog readability metric, and a lexical diversity measure\footnote{For extracting these features we use \\ \url{https://bitbucket.org/aswathyve/linguistic\_style\_features}.}.
From the LIWC lexicon, we select categories that we hypothesize to be relevant to expressions of expectation in health-related communication. 
As shown in \Cref{tab:liwc-feature-groups}, these categories are grouped into seven interpretable dimensions:
\emph{Uncertainty} (hedges and cognitive stance terms),
\emph{Time focus} (past, present, and future orientation),
\emph{Emotion} (positive and negative affect),
\emph{Social references} (self- and other-related pronouns),
\emph{Motivation} (goal- and need-related words),
\emph{Health} (illness and wellness vocabulary), and
\emph{Style} (readability and lexical diversity). 

To test for linguistic differences between expectation and non-expectation posts, we use the Mann–Whitney U test, a non-parametric alternative to the two-sample $t$-test that does not assume normality. We report effect sizes using Cliff's $\delta$, indicating the direction and magnitude of group differences, and adjust all $p$-values for multiple comparisons using the Benjamini--Hochberg FDR procedure~\citep{benjamini1995controlling}.
This procedure is conducted both at the group level (aggregated dimensions) and at the individual-feature level to identify which specific LIWC categories drive the observed effects.

\begin{table}
	\centering
	\small
	\begin{tabular}{p{1.5cm} p{5cm}} 
		\hline
		\textbf{Dimension} & \textbf{Features} \\
		\hline
		Uncertainty & tentat, certitude, cogproc, discrep, insight \\
		Time focus & focuspast, focuspresent, focusfuture \\
		Emotion & Affect, tone\_pos, tone\_neg, emo\_pos, emo\_neg, emo\_anx, emo\_anger, emo\_sad \\
		Social ref. & self\_reference, ppron, i, we, you, Social, family, friend \\
		Motivation & Drives, achieve, power, affiliation, need, want, lack, fulfill, reward, risk \\
		Health & health, illness, wellness, mental, substances \\
		Style & WPS, lexical\_diversity, gunning\_fog \\
		\hline
	\end{tabular}
	\caption{LIWC feature groups used in the analysis.}
	\label{tab:liwc-feature-groups}
\end{table}

\begin{figure*}
	\centering
	\includegraphics[width=1\linewidth]{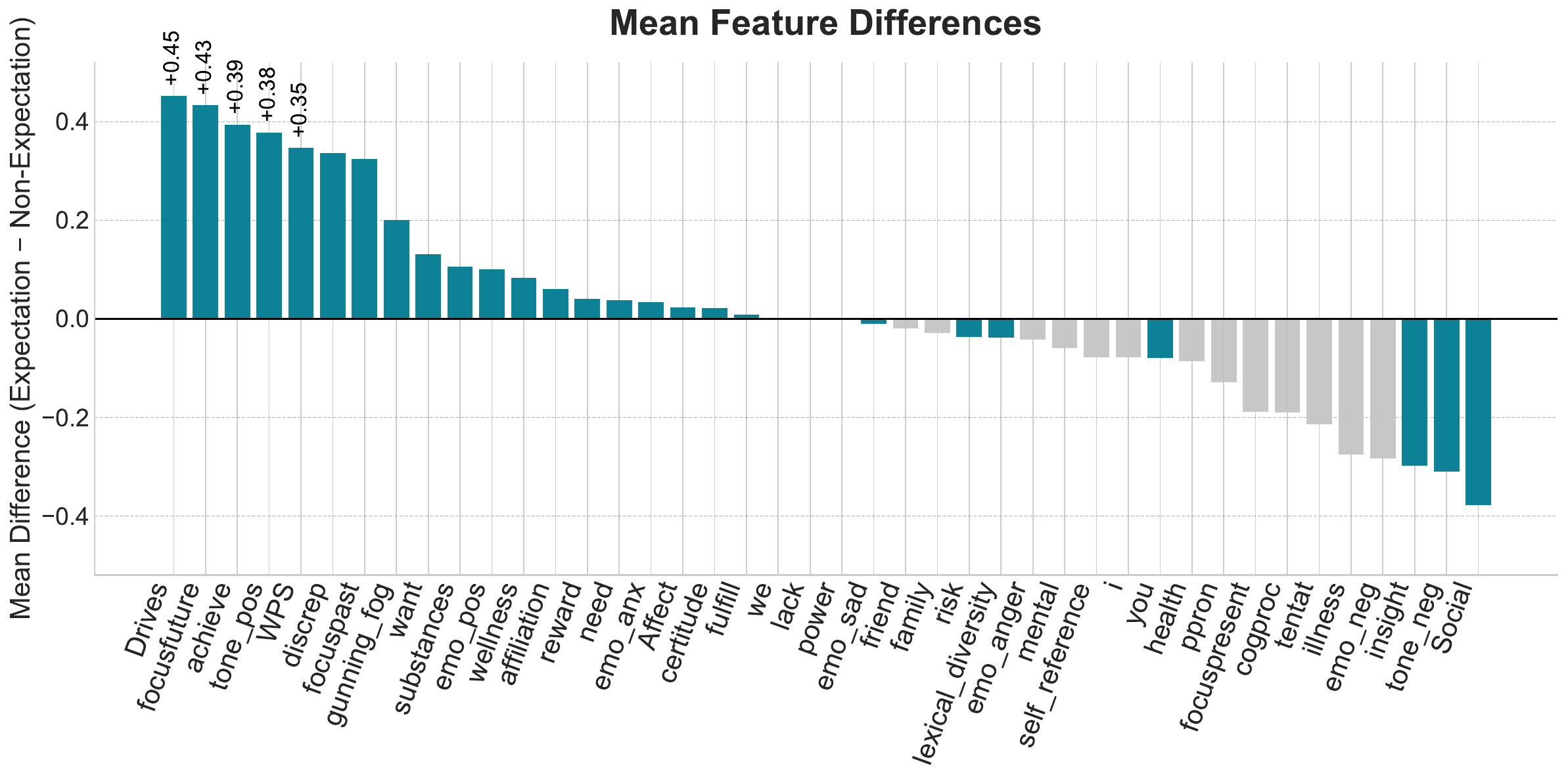}
	\caption{Mean feature difference in Expectation vs. Non-Expectations posts, with non-significant differences in grey.}
	\label{fig:liwcfeaturedifferences}
\end{figure*}

\subsection{Results}\label{sec:results_5}
As shown in \Cref{tab:length-stats}, within this dataset, posts labeled as containing expectations 
exhibit higher token counts (median = 138) than those without expectations (median = 102). The 
Mann--Whitney test confirms that this difference is statistically significant. While this finding 
does not imply a general relationship, it suggests that expectation-related content may be 
accompanied by more elaborate discourse in this sample.

As shown in \Cref{tab:liwc-groups}, expectation-related posts differ systematically from other 
health narratives across several linguistic dimensions.
The strongest effect is observed for \textbf{Motivation} ($\Delta=+0.11$), indicating greater 
use of goal-oriented and desire-expressing language. \textbf{Time focus} also shows a large overall 
difference ($\Delta=+0.19$), suggesting that expectation-related posts contain more 
temporal references in general, reflecting stronger temporal framing of events. 
When examining individual features within this category, the largest contribution 
comes from \textit{focusfuture} ($\Delta=+0.43$), showing that people tend to talk more 
about future outcomes when expressing expectations.
\textbf{Style} features also differ reliably ($\Delta=+0.18$), with posts expressing expectation containing longer 
and more syntactically complex sentences.
Smaller yet significant effects appear for \textbf{Social references} ($\Delta=-0.10$) and \textbf{Health} vocabulary ($\Delta=-0.05$), 
indicating fewer personal pronouns and health-specific terms in expectation posts.
Differences for \textbf{Emotion} and \textbf{Uncertainty} were not statistically significant.

\begin{table}
	\centering
	\small
	\begin{tabular}{lrrr}
		\hline
		\textbf{Dimension} & \textbf{$\Delta$(Exp–Non)} & \textbf{Cliff's $\delta$} & \textbf{FDR-$p$} \\
		\hline
		Motivation     &  +0.1084 &  0.1780 & $<\!10^{-41}$  \\
		Time focus     &  +0.1898 &  0.0912 & $<\!10^{-11}$ \\
		Style          &  +0.1800 &  0.0867 & $<\!10^{-9}$\\
		Social refs.   &  -0.0985 & -0.0389 & 0.0045 \\
		Health         &  -0.0465 &  0.0273 & 0.0482 \\
		Emotion        &  -0.0138 &  0.0229 & 0.0895 \\
		Uncertainty    &  -0.0689 & -0.0146 & 0.2581 \\
		\hline
	\end{tabular}
	\caption{Group-level LIWC differences (Mann–Whitney with Benjamini--Hochberg FDR). $\Delta$ is mean(Exp) $-$ mean(Non); positive values indicate higher means in expectation posts.}
	\label{tab:liwc-groups}
\end{table}

At the individual-feature level (Figure~\ref{fig:liwcfeaturedifferences}), 
expectation posts show increased use of words related to \textit{Drives}, \textit{achieve}, and \textit{tone\_pos}, 
highlighting a motivational and positive framing. 
Conversely, categories such as \textit{risk}, \textit{emo\_anger}, \textit{emo\_neg}, and \textit{tone\_neg} 
tend to have higher mean scores in non-expectation posts. 
Although not all of these differences reach statistical significance, the overall pattern suggests that 
expectation-related discourse is characterized by a more optimistic and action-oriented tone, 
whereas non-expectation posts more often convey caution, frustration, or negative affect.\

\begin{figure}
	\centering
	\includegraphics[width=1\linewidth]{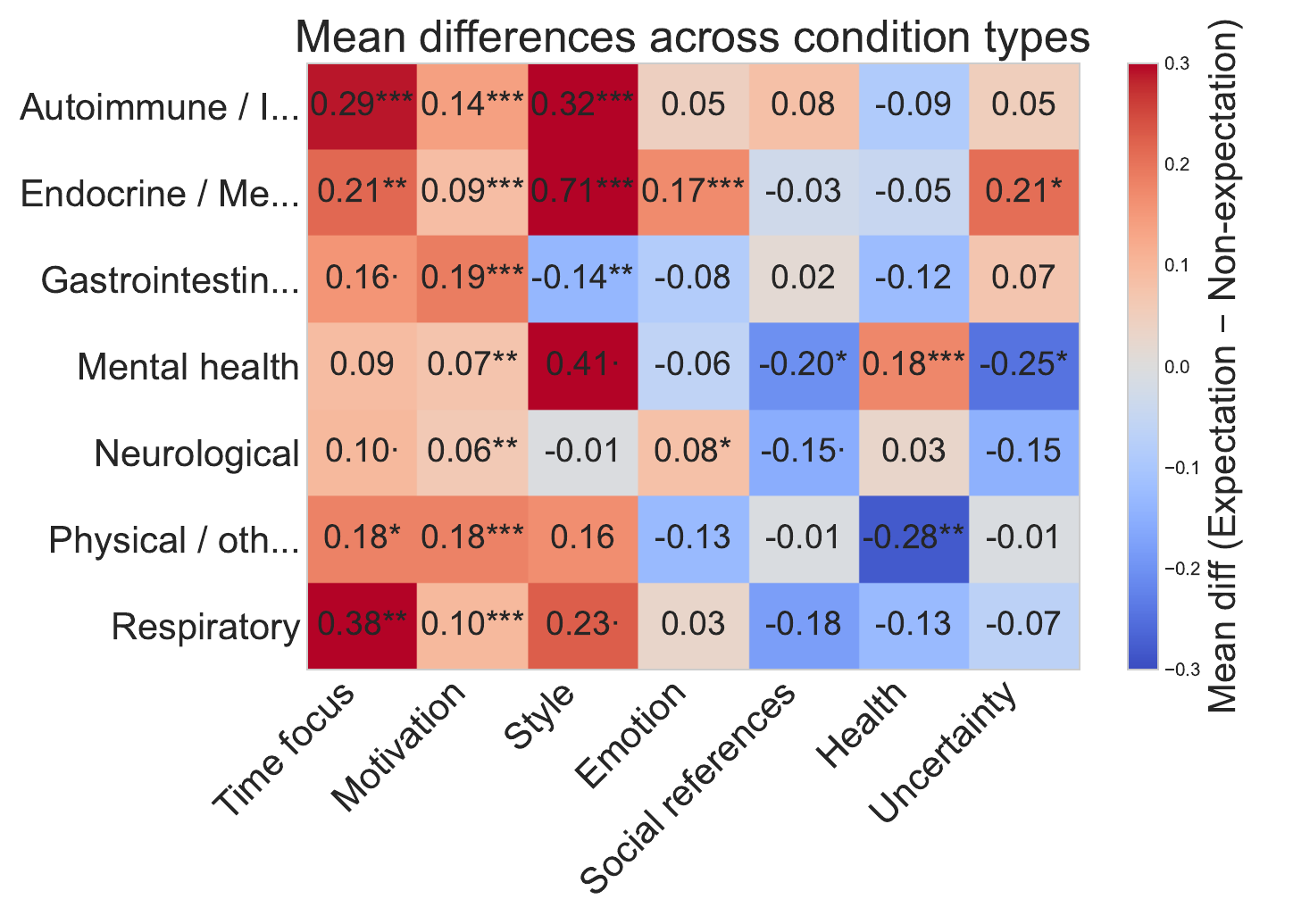}
	\caption{Mean differences by condition types.}
	\label{fig:diseasediff}
\end{figure}

To examine if the observed patterns hold across different medical conditions, we group the conditions into seven categories: 
\textit{Mental health}, \textit{Autoimmune/Inflammatory}, \textit{Endocrine/Metabolic}, \textit{Neurological}, \textit{Gastrointestinal}, \textit{Respiratory}, and \textit{Physical/Other}.  As shown in Figure \ref{fig:diseasediff}) across medical condition types, expectation-related posts consistently show a more future-oriented and motivational tone than non-expectation narratives. The effect is strongest in endocrine/metabolic and respiratory conditions, where posts were also longer and stylistically more complex, suggesting deliberate and goal-directed framing. In contrast, mental health discussions exhibited weaker differences and fewer social references, indicating that expectations in this context tend to be more introspective and self-focused. Overall, these patterns highlight that while the linguistic signature of expectation is broadly stable across domains, its expression varies with the nature of the health condition and the type of experience patients describe.

\section{What patients expect and what actually happens?}
\label{sec:rq2}

While the linguistic analysis in \Cref{sec:rq1} characterizes 
\textit{how} expectations are expressed at the surface level (lexical and stylistic features), this section focuses on \textit{what} patients expect and \textit{why}. Therefore, we conduct a targeted manual analysis of a representative subset of the corpus.

As discussed in \cref{subsec:human}, by manually annotating the 245 posts for validating the quality of annotations, we identified 389 Treatment--Expectation--Outcome triplets . In this section, we annotate each of these triplets with the \textbf{Expectation type} (\textsc{Benefit}, \textsc{Harm}, \textsc{Worsening}, \textsc{No effect}, \textsc{Mixed}, or \textsc{Other}), 
the \textbf{Expectation basis} (\textsc{Personal}, \textsc{Social}, \textsc{Authority}, \textsc{Information/Media}, \textsc{Cultural}, \textsc{Self-efficacy}, or \textsc{Other}), 
and the \textbf{Outcome basis} (\textsc{Benefit}, \textsc{Harm}, \textsc{Worsening}, \textsc{No effect}, \textsc{Mixed}, or \textsc{Other}). Together, these annotations provide a preliminary understanding of how treatment expectations are formed, what factors motivate them, and how they relate to the outcomes that patients report.

\begin{figure}
	\centering
	\includegraphics[width=1\linewidth]{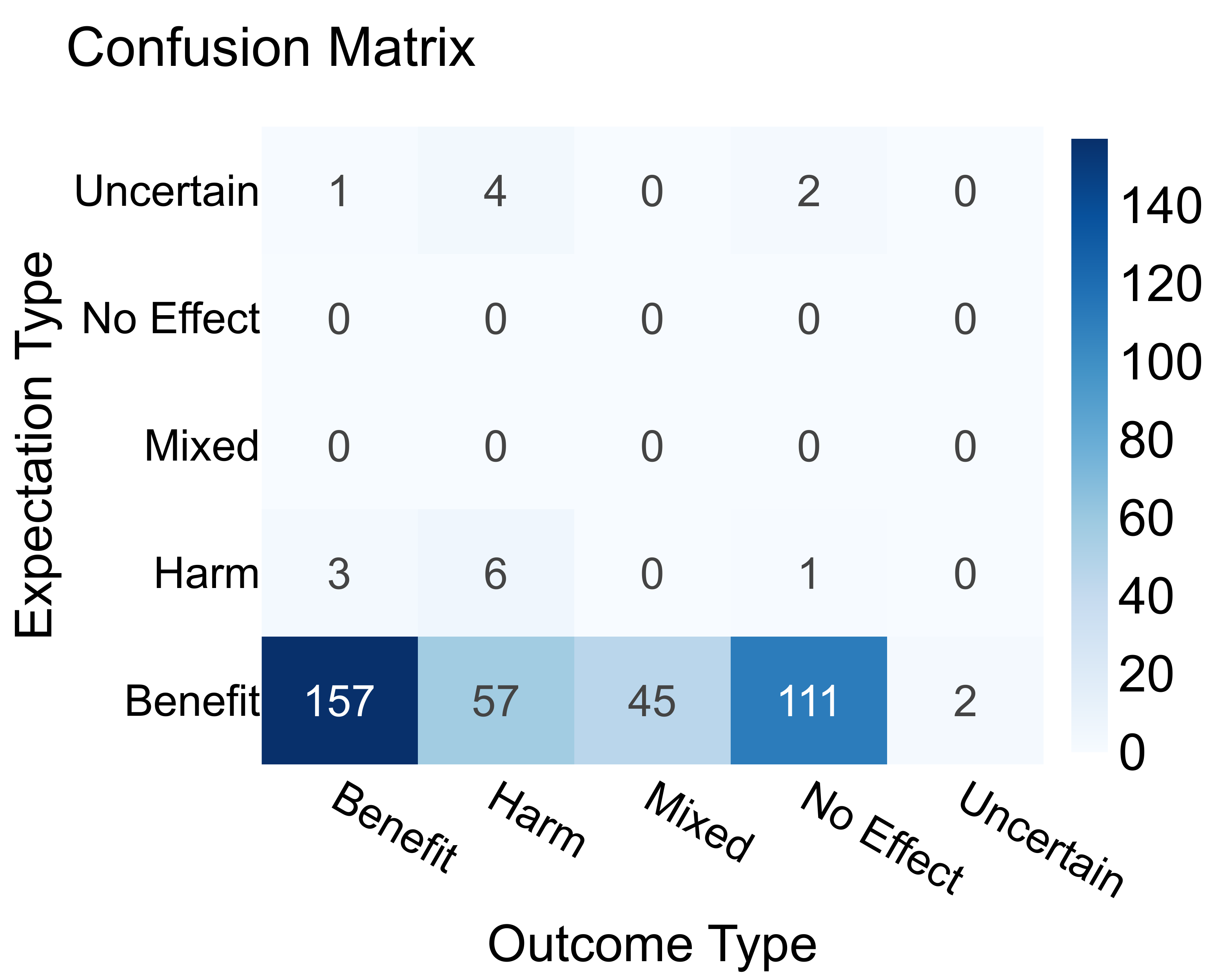}
	\caption{Confusion matrix representing alignment of expectation and outcomes types.}
	\label{fig:confusionmatrix}
\end{figure}

\begin{figure*}
	\centering
	\includegraphics[width=1\linewidth]{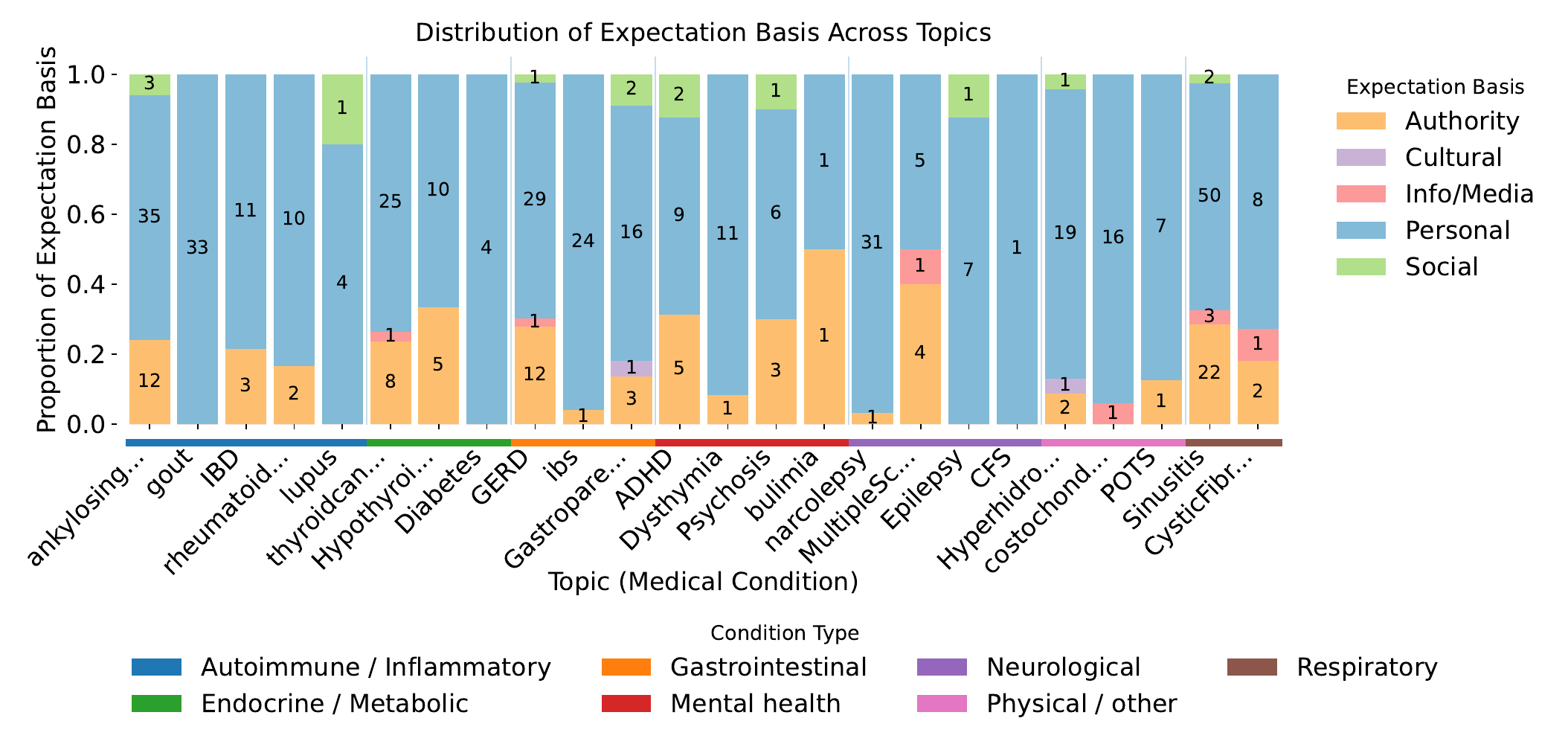}
	\caption{Distribution of expectation basis across medical conditions grouped by condition types.}
	\label{fig:basis}
\end{figure*}

\paragraph{Expectation Types.} Within the manually verified triplets, expectations are overwhelmingly \emph{benefit}-oriented: 372 out of 389 (95.6\%) expectations are labeled as \textsc{Benefit}, compared to \textsc{Harm} (10; 2.6\%) and \textsc{Uncertain} (7; 1.8\%); no instances were annotated as \textsc{Mixed} or \textsc{No Effect} at the expectation level in this sample. In line with the observation made in \Cref{sec:results_5}, we see that in this limited sample, people predominantly articulate \emph{benefit}-oriented expectations.

\paragraph{Expectation--Outcome alignment.}

Figure~\ref{fig:confusionmatrix} shows the alignment between the treatment
expectations and the actual outcomes reported by the patients. As shown in Figure~\ref{fig:confusionmatrix}, most \textsc{Benefit} expectations are followed by \textsc{Benefit} outcomes (157/372; 42.2\%), yet nearly one third correspond to \textsc{No Effect} (111/372; 29.8\%) and a smaller share to \textsc{Harm} (57/372; 15.3\%) or \textsc{Mixed} (45/372; 12.1\%). The few \textsc{Harm} expectations show some alignment with harmful outcomes (6/10; 60.0\%), and \textsc{Uncertain} expectations tend to accompany neutral or negative outcomes (4/7 and 2/7, respectively). Since majority of the sampled instances contain expectation if \textsc{Benefit}, it is hard to draw a strong conclusion. A larger-scale analyses is necessary to test whether expectations systematically align with outcomes in naturally occurring patient discourse. 

\paragraph{Linguistic certainty and expectation strength.}
The strength or confidence in their expectations ("I think" vs. "I am absolutely certain") may also influence the corresponding treatment outcomes. 
To approximate this dimension, we derive a \textit{Certainty score} from two LIWC features that capture epistemic stance: \texttt{certitude} and \texttt{tentat}. 
\[
\text{Certainty score} = z_{\text{certitude}} - z_{\text{tentat}} 
\]
where, \texttt{certitude} contributes positively, reflecting confident or assertive phrasing, while \texttt{tentat} contributes negatively, capturing hedging or uncertainty.  Higher Certainty score indicates more confident or decisive language, whereas lower value reflects tentative or reflective expression.
We then compare this score between \textit{aligned}(Expectation type = Outcome type) and \textit{misaligned}(Expectation type $\neq$ Outcome type) posts using a Mann–Whitney U test to assess distributional differences.

We observe a mean index close to zero for both aligned ($M=-0.027$) and misaligned ($M=0.020$) instances, and the difference was not statistically significant (Mann–Whitney $p=0.86$; Cliff's $\delta=-0.011$). These results suggest that linguistic certainty, as captured by LIWC features is not predictive of expectation-outcome alignment. 

Overall, this exploratory finding highlights that certainty of expectations may not be easily captured by lexicon-based measures. Future work should therefore include explicit certainty annotations or context-sensitive modeling (e.g., modality or confidence detection) to better capture how strongly patients believe in their stated expectations.

\paragraph{Expectation basis across health topics.}
Figure~\ref{fig:basis} illustrates how the rationale behind patients' expectations varies across different medical conditions. Across all topics, expectations are predominantly grounded in \textit{personal} experience, reflecting that most users draw on their own prior treatment history or symptom when anticipating outcomes. References to \textit{authority} (e.g., medical professionals or prescriptions) constitute the second most common basis and appear more prominently in certain conditions such as Sinusitis, GERD, and Ankylosing Spondylitis, though without a consistent pattern across condition types. Mentions of social or \textit{info/media} sources occur infrequently, while \textit{cultural} references are rare in the current subset. Overall, this pattern suggests that patients' expectations in the current subset are shaped largely by individual experience and clinical authority rather than by broader social, informational, or cultural influences. 


\section{Conclusion}
\label{sec:conclusion}

In this paper, we introduce the novel task of expectation detection which aims to
capture how beliefs and anticipations about the future are expressed in language. To understand and formalize the notion of expectation, we conduct a case study on health communication,
where expectations about treatment outcomes are known to have an influence on the effectiveness of the treatment. To enable this task, we create \dataset, a corpus of
patient-authored Reddit posts expressing expectation and a subset of the dataset
annotated with Treatment-Expectation-Outcome(TEO) triplets. Our analyses show that posts with expectations are longer, more future-oriented and
motivational than other narratives. Manual verification of TEO triplets showed that 
expectations in the subset are mostly discussing benefits rather than negative outcomes.

Overall, this work takes a first step towards modeling expectation as a measurable linguistic phenomenon. With the qualitative analysis on the treatment expectations, basis and outcomes, we highlight the opportunities and challenges in computational research on expectation effects using large scale
social media data. In future work, we aim to examine how expectations vary in their temporal orientation and level of certainty.

\section*{Ethical Considerations}
The dataset used in this study builds on the publicly available \redhot corpus~\cite{wadhwa-etal-2023-redhot}, which was constructed in compliance with Reddit's terms of use and includes an opt-out procedure for users. We retrieved the data from the Reddit post identifiers provided in the original dataset and excluded instances that had been deleted by users. No personally identifying information was collected or shared, and we did not attempt to identify or contact any users. In line with the ethical principles established by the original authors, we do not redistribute raw post texts. Instead, we release the annotations, along with scripts needed to reproduce the dataset by retrieving the posts directly from Reddit\footnote{The dataset and script for downloading are available at \url{https://www.ims.uni-stuttgart.de/data/RedHOTExpect}}.

Modeling expectations in health-related discourse has potential benefits for understanding patient perspectives but also entails ethical responsibility. We therefore emphasize that all analyses are conducted at an aggregate level and are not intended to make inferences about individual users.

\section*{Limitations}
The present study offers an initial step toward modeling expectations in language but has several limitations. The corpus is derived entirely from Reddit, representing a specific subset of patients~\cite{duh-et-al_2016} and a limited range of health-related subreddits; thus, the findings may not generalize to other populations, platforms, or clinical contexts. The results should therefore be interpreted with the understanding that this is a case study, and further investigation is needed to draw more robust conclusions about the linguistic properties of expectation. 

Further, the Treatment–Expectation–Outcome annotations were generated automatically using an LLM, and although manual validation indicated good accuracy, the data inevitably contain noise and inconsistencies. Moreover, LLM-based annotations are sensitive to prompt design and model behavior, which may introduce systematic biases that are difficult to control.

\section*{Acknowledgements}
This work is part of the SoftwareCampus project \textsc{Placebo}, which is funded by German Federal Ministry of Research, Technology and Space (BMFTR) under the grant 01IS23072.

\newpage
\section*{References}\label{sec:reference}

\bibliographystyle{lrec2026-natbib}
\bibliography{literature}
\clearpage
\appendix
\onecolumn

\section{Appendix}

\subsection{Prompt Design}\label{sec:appendix-prompt}
In this section, we provide the prompts used for our annotation pipeline. We employ two types of prompts: (1) a binary classification prompt for detecting the presence of Expectation Events (EEs) , and (2) a structured extraction prompt for identifying Treatment–Expectation–Outcome (TEO) triplets. Both prompts are designed to guide the model toward consistent and task-specific outputs, with clearly defined instructions, constraints, and output formats. The expectation detection prompt targets the identification of outcome-related expectations in text. The TEC prompt focuses on extracting fine-grained relational information between treatments, expectations, and outcomes. The full prompts are provided below for reproducibility.

\subsubsection{Expectation Detection Prompt}

\begin{quote}
\textbf{System:}

\begin{promptblock}
You are an expert annotator. Your task is to determine whether the text expresses an "Expectation Event (EE)."

Definition of Expectation Event (EE):  
An EE is a statement or implication about a possible or likely outcome (benefit, harm, no change, or worsening of condition) tied to a treatment, medication, or care action. 
Expectations may be prospective (future-oriented, concerning possible or likely outcomes such as benefit, harm, no change, or worsening) or retrospective (recalling a past belief about what was expected, in relation to the outcome that actually occurred).

Important:  
- The expectation must concern the **treatment outcome**, not a possible diagnosis, appointment, or unrelated event.  
- Outcome-related expectations include:  
  - Benefit: expected symptom relief, functional improvement, disease control.  
  - Adverse Effects: expected side effects or emotional impact.  
  - No Effect: skepticism about benefit.  
  - Worsening: anticipated deterioration or progression.  
- Expectations can be:  
  - Prospective (future-oriented) or retrospective (recalling a past belief).  
  - Explicit (directly stated) or implicit (implied through goals/conditions).  
  - Positive, negative, neutral, or mixed.  

Here are some examples:  
Expectation about treatment outcome counts:  
   - ``I expect this new medication will reduce my pain within two weeks."  
   - ``If the surgery works, I can get back to work."  
   - ``I was worried the drug would make me nauseous." (retrospective)  
Non-outcome expectation does NOT count:  
   - ``I expect to see my doctor again next month for a check-up."  
   - ``I expect they will finally diagnose me with lupus."  

Your output should strictly be in JSON format:  
- {{"expectation": true}}  
- {{"expectation": false}}  

Respond ONLY with the JSON object, nothing else. Answer with very minimal reasoning. 
Avoid step-by-step explanations unless strictly necessary.
\end{promptblock}

\vspace{0.3cm}

\noindent\textbf{User:}

\begin{promptblock}
The given text is a Reddit post where a person shares their experience related to a medical condition or treatment.  
Determine whether the text expresses an Expectation Event (EE).  

Strictly respond ONLY with a JSON of format: {{"expectation": true}} or {{"expectation": false}}.  
Text: {post}
\end{promptblock}

\end{quote}

\subsubsection{TEC Prompt Design}

\begin{quote}
\textbf{System:}

\begin{promptblock}
You are an expert annotator. Extract all (Treatment, Expectation, Outcome) triplets (TEC) from a Reddit post and, if present, the author's own follow-up comments.

Definitions:
- Treatment: any medication, surgery, therapy, care action, or dosage/regimen change.
- Expectation: the author's belief/anticipation about what the treatment will do (benefit, harm, no effect), explicit or implicit. Return exact spans if explicitly stated; summarize if only implied.
- Outcome (observed_outcome): the realized/actual result reported after the treatment (improvement, worsening, side effect, or no change).

Rules:
1. UNIT OF ANNOTATION = Triplet
   - Each triplet must connect exactly ONE treatment to ONE expectation to ONE outcome (observed_outcome).
   - Never mix expectations/outcomes from different treatments.
   - Return exact spans if explicitly specified, summarize if implicitly specified, else return null.

2. EXPECTATION REQUIREMENT
   - Must be about treatment outcomes (benefit, harm, or no effect) that the patient is undergoing or had undergone.
   - Strictly exclude diagnosis expectations, appointments, or logistics.
   - Both explicit and implicit expectations count.
     - Explicit: ``Antibiotics will clear this up."
     - Implicit: ``Antibiotics didn't help" (implies expectation they should have helped).
   - If no expectation is expressed for a treatment, no triplet is created.

3. OBSERVED OUTCOME
   - observed_outcome = what actually happened after the treatment (symptom change or side effect).
   - If only an expectation is given and no result is mentioned, set observed_outcome = null.
   - If both an expectation and a statement about what happened are present, use that statement as observed_outcome.
   - If the author later provides the result in a follow-up comment, update the triplet with that observed_outcome.
   - IMPORTANT: Receiving or taking the treatment itself (e.g., ``got the flu shot", ``took the pill") is NOT a treatment related outcome. 

4. TREATMENT REQUIREMENT
   - Must be a concrete intervention (drug, surgery, therapy, self-care, dosage/regimen change).
   - Vague ``it/this" counts only if clearly referring to a treatment.
   - If expectation/outcome exists without a treatment, set treatment = null.

5. NULL HANDLING (STRICT)
   - If one component is missing, set it explicitly to null.
   - Example: ``Let's see if it works" -> treatment = null, expectation = ``see if it works", observed_outcome = null.

6. COMMENTS
   - Look into the author's own follow-up comments for additional expectations/outcomes.
   - Ignore generic conversational replies (e.g., ``This is miserable", ``Did you get surgery?").
   - If later comments update the outcome for the same treatment/regimen, merge into the original triplet (do not duplicate).
   - If a comment introduces a new treatment/regimen, create a separate triplet.

7. MULTIPLE TRIPLETS
   - If multiple treatments/expectations/outcomes appear, output all of them separately.
   - Do not merge across different treatments.

8. IF NOTHING FOUND
   - If no valid triplets are detected, return:
     { "triplets": [] }

Output format:
{ "triplets": [ { "treatment": "...", "expectation": "...", "observed_outcome": "..." } ] }

Examples:

Text: "Doctor gave me antibiotics, I hope they clear my infection. Using a nasal spray too but it doesn't help my congestion."
-> { "triplets": [
     { "treatment": "antibiotics", "expectation": "hope they clear my infection", "observed_outcome": null },
     { "treatment": "nasal spray", "expectation": "should help with congestion", "observed_outcome": "no improvement" }
   ] }

Text: "Started melatonin, hoping I'll sleep better."
Comment: "Update: I'm sleeping through the night now."
-> { "triplets": [
     { "treatment": "melatonin", "expectation": "hoping I'll sleep better", "observed_outcome": "sleeping through the night" }
   ] }

Text: "Zoloft 25mg didn't do much. Upped to 50mg; mood is finally better."
-> { "triplets": [
     { "treatment": "Zoloft 25mg", "expectation": "should help with mood", "observed_outcome": "no improvement" },
     { "treatment": "Zoloft 50mg", "expectation": "should help with mood", "observed_outcome": "mood improvement" }
   ] }

Text: "I thought it wouldn't work, but it did. Scheduled to see ENT next week."
-> { "triplets": [
     { "treatment": null, "expectation": "thought it wouldn't work", "observed_outcome": "improvement" }
   ] }

Use null for any missing field.
Respond ONLY with the JSON object, nothing else. 
Avoid repeating the post. Keep reasoning to very low. Strictly avoid step by step reasoning.
\end{promptblock}

\vspace{0.3cm}

\noindent\textbf{User:}

\begin{promptblock}
Extract all (Treatment, Expectation, Outcome) triplets from the Reddit post and author comments.
If any component is missing, set it explicitly to null.
Respond ONLY with the JSON object, following this format exactly:
{{ "triplets": [ {{ "treatment": "...", "expectation": "...", "observed_outcome": "..." }} ] }}
Avoid repeating the post. Keep reasoning to very low. Strictly avoid step by step reasoning.

POST: {post}
COMMENTS: {author_comments}
\end{promptblock}

\end{quote}

\end{document}